\pdfoutput=1
\documentclass[11pt]{article}

\usepackage[preprint]{acl}

\usepackage{times}
\usepackage{latexsym}

\usepackage[T1]{fontenc}
\usepackage[utf8]{inputenc}

\usepackage{microtype}

\usepackage{inconsolata}

\usepackage{graphicx}

\usepackage{amsfonts}
\usepackage{booktabs}
\usepackage{siunitx}
\usepackage{graphicx}
\usepackage{soul}

\title{RingFormer: Rethinking Recurrent Transformer with Adaptive Level Signals}

\author{
    Jaemu Heo\thanks{\ \ Equal contribution}, \quad
    Eldor Fozilov\footnotemark[1], \quad
    Hyunmin Song, \quad
    Taehwan Kim \\
    IMML Lab, UNIST \\
    \texttt{\{skek000,eldorfozilov,hyunminsong,taehwankim\}@unist.ac.kr}
}

\begin{document}
\maketitle
\begin{abstract}
   Transformers have achieved great success in effectively processing sequential data such as text. Their architecture consisting of several attention and feedforward blocks can model relations between elements of a sequence in parallel manner, which makes them very efficient to train and effective in sequence modeling. Even though they have shown strong performance in processing sequential data, the size of their parameters is considerably larger when compared to other architectures such as RNN and CNN based models. Therefore, several approaches have explored parameter sharing and recurrence in Transformer models to address their computational demands. However, such methods struggle to maintain high performance compared to the original Transformer model. To address this challenge, we propose our novel approach, \emph{RingFormer}, which employs one Transformer layer that processes input repeatedly in a circular, ring-like manner, while utilizing low-rank matrices to generate input-dependent level signals. This allows us to reduce the model parameters substantially while maintaining high performance in a variety of tasks such as translation and image classification, as validated in the experiments.
\end{abstract}

\section{Introduction}
\label{intro}

Transformer models, since their introduction \cite{vaswani2017attention}, have dramatically transformed the landscape of deep learning, particularly excelling in tasks involving sequential data such as natural language processing \cite{brown2020language, radford2019language} and machine translation \cite{ott2018scaling}. Not long after their inception, they have also shown strong performance in various other domains such as reinforcement learning \cite{chen2021decisiontransformer}, image classification \cite{dehghani2023scaling, dosovitskiy2020image, liu2021swin}, object detection \cite{carion2020detr} and image generation \cite{jiang2021transgan, Peebles2022DiT, zhang2022styleswin}. Their core architecture, characterized by self-attention mechanisms and feedforward neural networks, enables effective handling of long-range dependencies and parallel processing of input sequences. The ability of this architecture to model intricate relationships within data has led to significant breakthroughs, making it a foundation model across many modern large-scale AI systems \cite{claude, google2023gemini, OpenAI_Achiam_Adler_Agarwal_Ahmad_Akkaya_Aleman_Almeida_Altenschmidt_Altman_etal._2024, touvron2023llama2}.

However, the impressive capabilities of transformer models come with substantial computational and memory costs \cite{brown2020language, dosovitskiy2020image}. The standard Transformer architecture consists of multiple layers, each containing millions of parameters that need to be trained and stored. This results in high memory usage and significant computational demands, often requiring specialized hardware. Moreover, deploying these models in resource-constrained environments, such as mobile devices or edge computing scenarios, becomes challenging due to their size and complexity. These limitations have spurred a growing interest in developing more parameter-efficient Transformer architectures \cite{dehghani2019universal, pires2023wide} that can retain their powerful performance while being more accessible and less resource intensive.

In this paper, we introduce a Transformer architecture that recurrently leverages a single shared Transformer block in a novel way by integrating input-dependent level signals at each block iteration, which are shown to be crucial for adapting the shared block to different stages of the model. The level signals are generated by depth-specific low-rank transformations applied to the input in the attention and feedforward layers within the Transformer block. Our \emph{RingFormer} model can also be viewed as stacking Transformer layers whose parameters combine (1) a set of global parameters shared across all Transformer layers and (2) a set of local low-rank layer-dependent parameters. This simple design effectively addresses the trade-off between reducing the number of model parameters and limiting the model's capacity to capture complex patterns.

We validate our model through experiments and analysis on machine translation and image classification. The results of experiments and analysis demonstrate that our model closely replicates the behavior of the original Transformer model, and it performs better against existing parameter-matched recurrence-based Transformer models, underscoring the effectiveness of our approach in maintaining high performance with fewer parameters.

The contributions of this paper are summarized as follows:
\begin{itemize}
    \item We enhance a recurrent Transformer architecture to significantly reduce the model's parameter count while maintaining high performance.
    \item We propose novel input-dependent level signals generated in a parameter-efficient way using low-rank matrices to improve the adaptability of a recurrent Transformer model, and show that those signals help the model replicate the behavior of the original model.
    \item We demonstrate the validity of our approach through careful analysis and ablation studies, and show the effectiveness of our model on tasks such as translation and image classification.
\end{itemize}

\section{Background}
\label{background}

\subsection{Transformer Architecture}
\label{2.1.transformer achitecture}
The Transformer architecture \cite{vaswani2017attention} comprises multiple layers of the same structure stacked together, with each layer consisting of two main modules: \emph{Attention} and \emph{Feedforward Network} described in Equations (\ref{eq_attn}) and (\ref{eq_ffn}), respectively. Each of these modules is accompanied by residual connections and layer normalization. In addition, to provide information about the position of tokens in the sequence, the Transformer model adds static sinusoidal or learnable positional encodings to the input embeddings. These encodings allow the model to capture the order within a sequence. The following equations describe the mechanism of two main modules:
\begin{equation}
\label{eq_attn}
    Attention(Q, K, V) = softmax(\frac{QK^T}{\sqrt{d}})V
\end{equation}
\begin{equation}
\label{eq_ffn}
    FFN(x) = \sigma(xW_{up}+b_{up})W_{down}+b_{down}
\end{equation} \\ [2pt]
Here, \( Q \), \( K \), and \( V \) are the results of projecting the input vectors through their respective matrices. Attention module can be classified into self-attention (when the \( Q \), \( K \), and \( V \) input vectors are the same) and cross-attention (when the \( Q \) input vector is different from the \( K \) and \( V \) input vectors), while the feedforward block consists of up-projection and down-projection transformations with non-linearity function $\sigma$ between them.

It is well known that Transformer architecture follows a scaling law for both vision tasks and NLP tasks \cite{dehghani2023scaling, hoffmann2022empirical}. This scaling law demonstrates that the performance of Transformer models improves predictably as the model size and computational resources increase. Due to the steep slope of the scaling law, the parameter sizes of Transformer models have continued to grow, leading to significant advancements in their capabilities. However, this growth has also made training and using such massive models increasingly infeasible without substantial GPU resources. 

\subsection{Related Work}
\label{2.2.pe transformer}
To address the challenge of requiring extensive hardware resources for large Transformer models, researchers have explored various methods to enhance efficiency.

One approach is related to pruning of Transformer model layers, which involves removing less important layers or weights to streamline the model. It was found that many deep layers in large language models are redundant \cite{gromov2024unreasonable}, and by pruning up to half of these layers, it was possible to significantly reduce the model size with minimal accuracy degradation. 

Another strategy is sharing parameters across different layers or components in Transformers, reducing the model's complexity and memory usage. The Universal Transformers \cite{dehghani2019universal} introduces a model where parameters are shared across layers using a recurrent mechanism with layer-dependent positional encoding, which maintains good performance in various NLP tasks while reducing the number of parameters. People have also proposed sequence and cycle strategies for sharing parameters across layers~\cite{takase2021lessons}, improving efficiency and performance in tasks like machine translation and speech recognition. Similarly, Subformer \cite{reid2021subformer} and One Wide Feedforward \cite{pires2023wide} investigate partial weight sharing within layers, showing that significant parameter reductions can be achieved with little accuracy sacrifice. These models demonstrate that shared parameters can lead to efficient and effective Transformer architectures.

To investigate recurrence-based models, we performed a layer representation similarity analysis using the common CKA (centered kernel alignment) \cite{kornblith2019similarity} method and mean attention distance (MAD) \cite{dosovitskiy2020image} analysis, and we found that the layer representations and internal attention behavior of the previously proposed fully recurrence-based Transformer model \cite{dehghani2019universal} are considerably different compared to those of the original Transformer model.

We hypothesized that the difference in model behavior, especially in attention module, might be the main cause for the gap in performance, and if we can simulate the behavior of the original model using a recurrent model with adaptive level signals, we can also maintain higher performance. Our proposed methodology is focused on addressing this difference, narrowing the gap of the model behavior, and in turn the model performance.

\section{Method}
\label{method}
\subsection{Overview}
In this section, we provide a detailed explanation of our proposed work, covering the specific details about the structure of our model.

The encoder or decoder Transformer-based models consist of several layers with the same structure, where each layer is a combination of sub-layers such as attention and feedforward layers. Those models can be formulated in the following way:

\begin{equation}
\begin{array}{cl}
\label{eq:recurrsive}
&F(x)=f_{N}(f_{N-1}(... f_2(f_1(x))))\\\\
&=f(f(...f(f(x, p_1), p_2)), p_{N-1}), p_{N})
\end{array}
\end{equation}
where $N$, $F$, $f$, $x$ and $p_i$ denote the number of layers, entire encoder (or decoder), each encoder (or decoder) block, input and parameters of each $i^{th}$ layer, respectively. The general formulation of the recurrent Transformer model with level transition functions can be written as below:

\begin{equation}
\centering
\begin{array}{cl}
&F(x) = f_N(f_{N-1}(...f_1(x)))\\\\
&f_i(x) = f_r(x, g_i(x))
\end{array}
\end{equation} \\
where $f_r$ denotes the recurrent Transformer block and $g_i(x)$ represents a generic level transition function specific for each level. In Universal Transformers \cite{dehghani2019universal}, it was shown that using static spatio-temporal positional embeddings can serve as level transition functions for the recurrent Transformer layer and have good model performance. Specifically, in that work, level transition function $g_i(x)$ can be represented as $g_i(x) = x + l(i, x_p)$, where $l$ is a function that returns a positional embedding vector based on level depth $i$ and the position $x_p$ of the vector $x$, while the $i^{th}$ Transformer block function $f_i(x)$ can be represented as $f_i(x) = f_r(g_i(x))$.

Below, we describe our way of constructing and integrating level transition function $g_i(x)$ to generate adaptive level signals.

\subsection{Adaptive Level Signals}
\label{3.1.level signals}
To have effective transition between the levels when using recurrent Transformer block, we make $g_i(x)$ directly dependent on the input in the following way: $g_i(x) = M_i \cdot x$, where $M$ is a learnable transformation matrix. Since the main role of level signals is to nudge the input vectors in the right direction, which is an easier task compared to the main input transformation done by the recurrent layer, we hypothesize that making the $M$ matrix low-rank while keeping the recurrent layers at full-rank will let us have parameter-efficiency and high performance at the same time. We draw inspiration for such a low-rank matrix construction and its weight initialization from the parameter efficient fine-tuning (PEFT) technique, LoRA \cite{hu2021lora}, and decompose \( M_i \) into two low-dimensional matrices, \( A_i \) and \( B_i \) described in Equation \ref{eq:low rank}.
\begin{equation}
\label{eq:low rank}
M_i = A_i\cdot B_i^{T},\  A_i, B_i \in \mathbb{R}^{d\times r}\ \text{and}\ r \ll d
\end{equation}

\begin{figure*}[t!]
\centering
\includegraphics[width=0.8\linewidth]{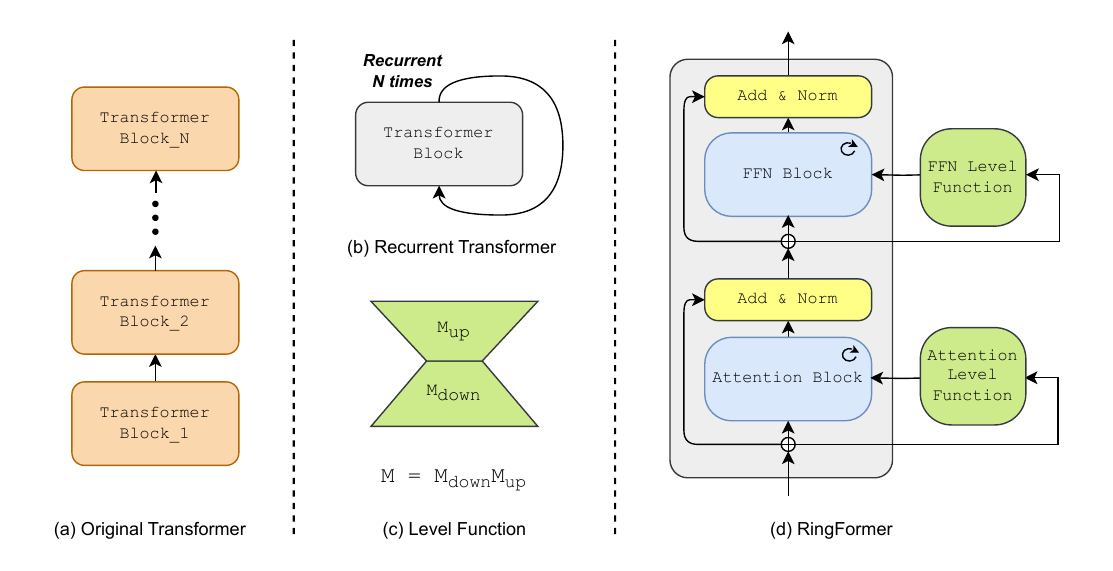}
\caption{Overview of (a) vanilla Transformer \cite{vaswani2017attention}, (b) recurrent Transformer \cite{dehghani2019universal} and (d) our RingFormer architecture. The Transformer block represents either encoder or decoder. In the RingFormer model, a single block consisting of attention and feedforward modules is used iteratively, with each sub-modules having unique layer normalization and level signals. (c) illustration of the low-rank matrices representing the level functions, where $M_{down}$ down-projects the input to a lower dimensional space and $M_{up}$ up-projects back to the original space.} 
\label{fig:model_01}
\end{figure*}

Since a Transformer layer consists of an attention block and a feedforward block, we generate two distinct signals $g_{Ai}(x)$ and $g_{Fi}(x)$: one for the attention block and the other for the feedforward block, respectively. Additionally, since the Transformer block also has layer normalization applied between the sub-layers, for each level, we allocate unique layer normalization in the attention and feedforward layers. This provides extra input adaptation while only slightly increasing the number of total parameters in the model.

\subsubsection{Attention Block}
For the attention mechanism, which calculates relevance between elements of a sequence using three projection matrices (query, key, and value), we generate level signals for each of those projections using separate low-rank matrices. We integrate signals after the projection of the input vector $x$ by $W_Q, W_K$ and $W_V$ matrices (shared across the levels) in the following way:

\begin{equation}
\label{eq.lev_attn_full}
\centering
\begin{array}{c}
Q_i = W_Q \cdot x + g_{A_{Qi}}(x),\; \\ \\
K_i = W_K \cdot x + g_{A_{Ki}}(x),\; \\ \\
V_i = W_V \cdot x + g_{A_{Vi}}(x), \\
\end{array}
\end{equation} where \space $g_{A_{Qi}}(x) = M_{Qi}\cdot x$, \space
$g_{A_{Ki}}(x) = M_{Ki}\cdot x$, \space $g_{A_{Vi}}(x) = M_{Vi} \cdot x$. By incorporating the level functions separately for Q, K, and V, we enable fine-grained control over depth-dependent modifications to each component of the attention mechanism. Also, adding the level signals in this manner avoids direct input changes to the main recurrent projections, which was found to be beneficial in our experiments. This can be because such a direct input change can interfere with the learning process of the recurrent layers in the attention module, which needs to solely focus on modeling effective communication between tokens.

\subsubsection{Feedforward Block}
For feedforward block, the projection of input to intermediate vector of this module requires relatively large number of parameters. Furthermore, there have been various explorations regarding the role of feedforward network in Transformers. One such study~\cite{geva2021transformer} argues that the feedforward network can be interpreted as a key-value memory pair, where the matrix of the first linear layer is involved in the coefficients of input factors, and the matrix of the second linear layer relates to information about the training corpus. Considering parameter-efficiency and the previous finding, in our approach, for the feedforward network, we add signals before projecting the input using the up-projection layer to guide the coefficient formation of the input in the following way:
\begin{equation}
\begin{array}{c}
FFN(x) = \sigma((x+g_{Fi}(x))W_{up})W_{down}
\end{array}
\end{equation}
where $g_{Fi}(x) = M_{Fi} \cdot x$, the function $\sigma$ is a non-linear function such as GELU \cite{hendrycks2023gaussian}, and the bias terms were omitted for brevity.

In encoder-decoder models, we iteratively reuse a single Transformer block consisting of attention and feedforward sub-blocks in the encoder, while the decoder utilizes a separate shared Transformer block with cross-attention, which is also shared across layers. For the recurrent cross-attention module inside the decoder, we do not incorporate level signals, as cross-attention takes on the output of the already level-adapted attention module as query and the representations from the encoder as key and value, which do not require additional adaptation at different levels. The overall structure of our model is illustrated in Figure \ref{fig:model_01}.

\section{Experiments}
\label{exp}

We evaluate the performance of our RingFormer model and baseline models across two tasks: machine translation and image classification.

For the baseline models, we choose the vanilla Transformer model \cite{dosovitskiy2020image, vaswani2017attention}, one recurrent transformer model, Universal Transformer \cite{dehghani2019universal}, and one partially recurrent model, One Wide Feed Forward model \cite{pires2023wide}, with specific adaptations to the corresponding tasks described below. 

\subsection{Experimental Details}
\label{4.1:training detail}
In this section, we provide detailed description of each downstream task to facilitate the reproduction of our experimental results. For all of our models, the rank of the decomposed matrices for the level signals is fixed at the input hidden dimension divided by 16. We perform ablations for the different ranks and show the results in Table \ref{table:ablation}.

\paragraph{Translation}
Transformer model is firstly proposed in translation task \cite{vaswani2017attention}. Thus, we also test our model on the translation task, with two model sizes shown in Table 1. We train all models on WMT-14 \cite{bojar-etal-2014-findings} German-English dataset which consists of 4.5M pairs of sentences. For evaluation, we calculate BLEU score \cite{papineni2002bleu} for the WMT-14 German-English test set and we employ BiBERT vocabulary with bi-lingual tokenizer with vocab size equal to 52K \cite{xu2021bert}. We set the number of layer (iteration), batch size and entire training step as 6, 512 and 830K for base and large setting on two A100 80GB GPUs, respectively. In the training session, we used Adam \cite{kingma2017adam} optimizer with a cosine learning rate scheduler having 40K of warm-up steps. Also, we used GELU \cite{hendrycks2023gaussian} as activation function for all models.

The main model hyperparameters and experiment results are given in Table \ref{table:Translation}, where we report the parameter size except those parameters in the encoder, decoder and vocabulary head because their count is the same for all models having the same hidden input dimension and feedforward block dimension. For base size models, the encoder and decoder embedding layer each consists of 26.62M parameters, and vocabulary head contains 26.67M parameters; for large size models, 53.24M for encoder and decoder embedding layer, 53.30M for vocabulary head. 

\begin{figure*}[h!]
    \centering
    \begin{minipage}{0.3\textwidth}
        \includegraphics[width=\textwidth]{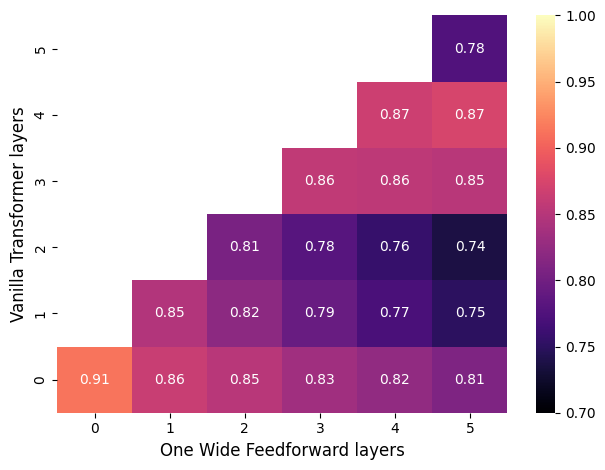}
    \end{minipage}%
    \hfill
    \begin{minipage}{0.3\textwidth}
        \includegraphics[width=\textwidth]{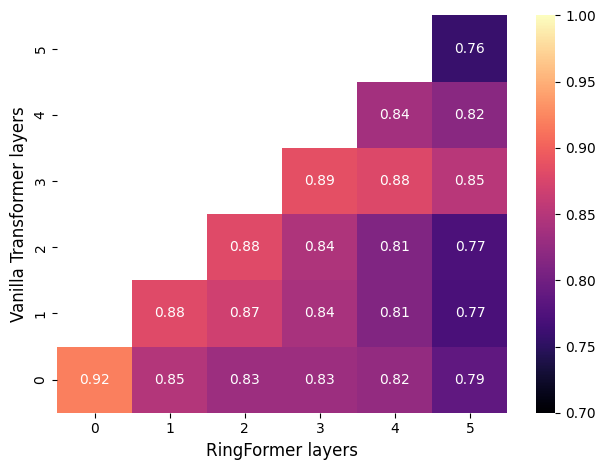}
    \end{minipage}%
    \hfill
    \begin{minipage}{0.3\textwidth}
        \includegraphics[width=\textwidth]{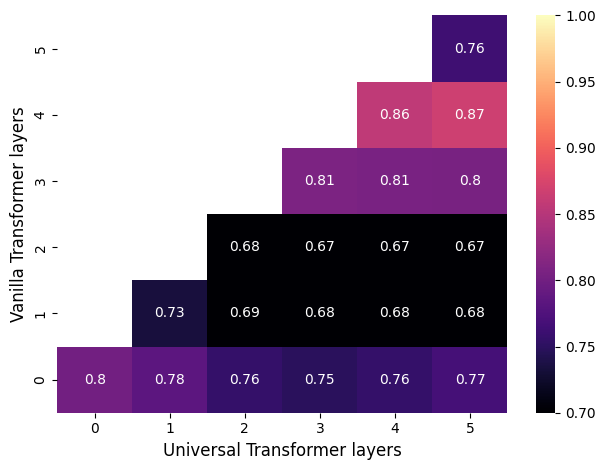}
    \end{minipage}%
    \vfill
    \begin{minipage}{0.3\textwidth}
        \includegraphics[width=\textwidth]{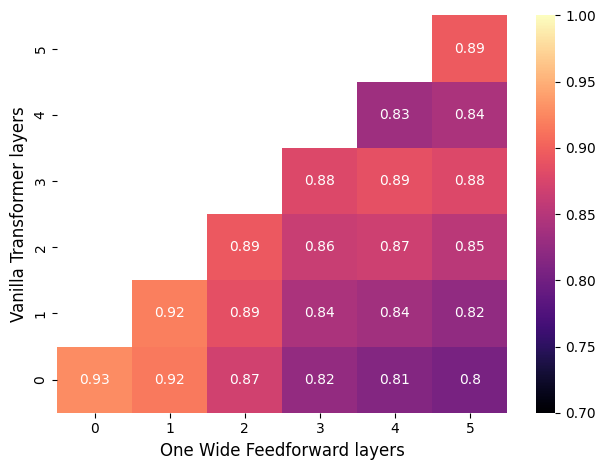}
    \end{minipage}
    \hfill
    \begin{minipage}{0.3\textwidth}
        \includegraphics[width=\textwidth]{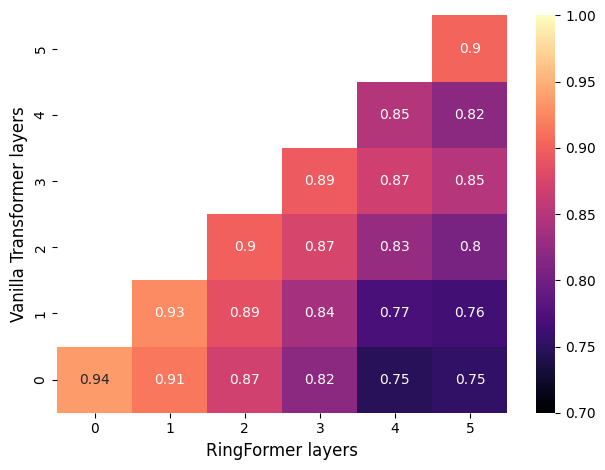}
    \end{minipage}
    \hfill
    \begin{minipage}{0.3\textwidth}
        \includegraphics[width=\textwidth]{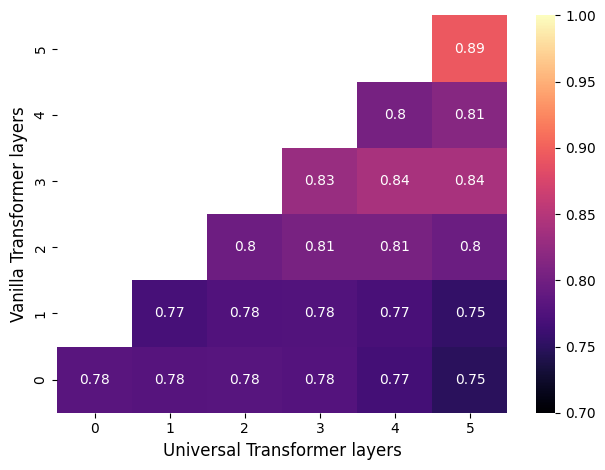}
    \end{minipage}
    \caption{Representation Similarity Analysis using CKA (centered kernel alignment) \cite{kornblith2019similarity} for the base-size models trained on the translation task. The figures on the upper row are for the encoder part. The figures on the lower row means are for the decoder part. All models have 6 number of layers / iterations. The values on the figures are between 0 and 1, where higher values indicate more similarity of layers between models.}
    \label{fig:grid_cka_base}
\end{figure*}

\paragraph{Image Classification}
As the ViT \cite{dosovitskiy2020image} model became very prevalent in the vision domain, especially in image classification, we decided to test our model and other baseline models on this task. The models are adjusted to have only encoder layers, which take image patches with a class token attached as an input, and perform the prediction using the hidden state of the class token from the last layer. For the Vision Transformer (ViT) model \cite{dosovitskiy2020image}, we stick to the original architecture, while for the Universal Transformer \cite{dehghani2019universal}, static sinusodial spatio-temporal positional embeddings are used as level transition function between the levels in the encoder. For the One Wide Feed Forward model \cite{pires2023wide}, the feedforward layer is shared across the levels, while the attention layer parameters are distinct for each level.

We first train smaller models on a subset of the original ImageNet-1K dataset \cite{deng2009imagenet} for 100 epochs. We randomly chose 100 classes with the total number of 100K training samples (1K per each class) from the original training set, and 5K testing samples (50 per each class) from the original validation set. For easy referencing, we call that subset \textit{ImageNet-small}. As the size of the dataset is relatively small, we decided to train models having only 6 layers / iterations (in the case of recurrent models, we say iterations or levels instead of layers). For bigger size models with 12 layers / iterations, we trained on the whole \textit{ImageNet-1K} for 50 epochs due to limited resources. 

The additional training and \textit{ImageNet-small} dataset details are given in Appendix  \ref{appendix:implementation_details} and \ref{appendix:imagenet_small_dataset}. The model hyperparameters, parameter size and experiment results on \textit{ImageNet-small} and \textit{ImageNet-1K} are given in Table \ref{table:vision_mini_imagenet} and \ref{table:vision_imagenet}.

\subsection{Experimental Results}

\begin{table}[h!]
\small
\setlength{\tabcolsep}{1.5mm} 
\centering
\begin{tabular}{c|c c|c}
\toprule
{$model$} & {$H / FF$} & {$P^*$} & {$BLEU\uparrow$} \\
\midrule
Vanilla Transformer & 512 / 2048 & 44.05M & \textbf{30.46} \\
One Wide FFN & 512 / 2048 & 20.98M & \underline{29.54} \\
Universal & 512 / 2048 & 7.34M & 29.12 \\
\textbf{RingFormer} & 512 / 2048 & 8.94M & 29.52 \\
\midrule
Vanilla Transformer & 1024 / 4096 & 176.18M & \textbf{30.96} \\
One Wide FFN & 1024 / 4096 & 83.91M & 29.88 \\
Universal & 1024 / 4096 & 29.37M & 29.47 \\
\textbf{RingFormer} & 1024 / 4096 & 35.71M & \underline{29.96} \\
\bottomrule
\end{tabular}
\caption{Translation results on WMT-14 De-En \cite{bojar-etal-2014-findings}. We evaluated models based on test dataset BLEU score \cite{papineni2002bleu}, which is rounded to the second decimal place. \textbf{Bolded} score indicates the highest performance, \underline{underlined} score indicates the second highest performance. The $H$, $FF$, and $P^*$ represent the hidden input dimension, feedforward block dimension, parameter size (except parameters of embedding layer in encoder, decoder and vocabulary head), respectively.}
\label{table:Translation}
\end{table}

\paragraph{Translation} 
The details of experimental results on translation are presented in Table \ref{table:Translation}. Our RingFormer model achieves competitive performance with Vanilla Transformer model \cite{vaswani2017attention} and One Wide FFN model \cite{pires2023wide} with less number of parameters for base and large size models. RingFormer outperforms Universal model \cite{dehghani2019universal}, while having similar parameter size. These results also imply that our design choice for level-signals is more effective than adding input-independent sinusoidal vectors.

\begin{table}[h!]
\centering
\small
\begin{tabular}{c|c c|c} \toprule
    {$model$} & {$H / FF$} & {$P$} & {$Acc\uparrow$} \\ \midrule
    \text{ViT}              & \text{512} / \text{2048} & \text{19.36M} & \textbf{63.66\%} \\
    \text{UiT}            & \text{512}  / \text{2048} & \text{3.60M} & \text{58.64\%} \\
    \text{OWF$^d$} & \text{376}  / \text{1024} & \text{4.51M} & \text{58.62\%} \\
    \textbf{RingFormer}       & \text{512} / \text{2048} & \text{4.4M} & \underline{60.66\%} \\ \midrule
    \text{ViT$^{d}$}   & \text{328} / \text{1536} & \text{8.94M} & \underline{62.22\%} \\
    \text{UiT$^{s}$}   & \text{848} / \text{3072} & \text{8.84M} & \text{59.38\%} \\
    \text{OWF}              & \text{512} / \text{2048} & \text{8.86M} & \text{61.50\%} \\ 
    \textbf{RingFormer$^{s}$}       & \text{728} / \text{3072} & \text{8.82M} & \textbf{62.58\%} \\
    \bottomrule
\end{tabular}
\caption{Image classification results on \textit{ImageNet-small} (the subset of ImageNet-1K \cite{deng2009imagenet}). \textbf{Bolded} score indicates the highest performance, \underline{underlined} score indicates the second highest performance. The superscripts "d" and "s" represent that the models are downscaled and upscaled, respectively. The $H$, $FF$, and $P$ represent the hidden input dimension, feedforward block dimension, and total parameter size, respectively. The values for $P$ and $Acc$ were rounded to the second decimal place.}
\label{table:vision_mini_imagenet}
\end{table}

\begin{table}[h!]
\centering
\small
\begin{tabular}{c|c c|c} \toprule
    {$model$} & {$H / FF$} & {$P$} & {$Acc\uparrow$} \\ \midrule
    %\text{ViT} & \text{768} & \text{3072} & \text{86.42M} & \text{65.65\%} \\
    \text{ViT} & \text{768} / \text{3072} & \text{86.42M} & \textbf{65.65\%} \\
    \text{OWF} & \text{768}  / \text{3072} & \text{34.45M}
    & \underline{64.31\%} \\
    \text{UiT}    & \text{768} / \text{3072} & \text{8.45M} & \text{61.63\%} \\
    %\text{OWF} & \text{768}  & \text{3072} & \text{34.45M}
    %& \text{64.31\%} \\
    \textbf{RingFormer}       & \text{768} / \text{3072} & \text{12.02M} & \text{63.68\%} \\ \midrule
    \text{UiT$^s$}           & \text{1560} / \text{6240} & \text{31.99M} & \text{63.30\%} \\
    \textbf{RingFormer$^s$} & \text{1284} / \text{5120} & \text{31.95M} & \textbf{65.91\%} \\
    \bottomrule
\end{tabular}
\caption{Image classification results on \textit{ImageNet-1K} \cite{deng2009imagenet}). \textbf{Bolded} score indicates the highest performance, \underline{underlined} score indicates the second highest performance. The superscript "s" represent that the models are upscaled. The $H$, $FF$, and $P$ represent the hidden input dimension, feedforward block dimension, total parameter size, respectively. The values for $P$ and $Acc$ were rounded to the second decimal place.}
\label{table:vision_imagenet}
\end{table}

\paragraph{Image Classification} The experimental results for image classification are shown in Table \ref{table:vision_mini_imagenet} and \ref{table:vision_imagenet}. 

Using \textit{ImageNet-small}, we conducted experiments on the ViT \cite{dosovitskiy2020image} model, downscaled One Wide FFN (OWF$^d$) \cite{pires2023wide}, UiT \cite{dehghani2019universal} and our RingFormer model. The results, presented in the upper half of Table  \ref{table:vision_mini_imagenet}, indicate that the ViT model achieves the highest accuracy, which is expected as it has more than four times the number of parameters compared to the other models. However, our RingFormer model has the second best performance, outperforming the other models of the same size. In the below half of Table \ref{table:vision_mini_imagenet}, where we scale all the models to the size of One Wide FFN model, our model shows the best performance, which shows the effectiveness of our approach.

We observed similar tendency when we trained bigger size models on the \textit{ImageNet-1K} dataset, for which the results are shown in Table \ref{table:vision_imagenet}. When comparing the models with the same input hidden dimension and feedforward block dimension, ViT model showed the best result, but when we upscaled our RingFormer model (RingFormer$^s$) to match the size of OWF model, it outperformed the two baseline models (OWF and UiT$^s$), and also showed slightly higher performance compared to the ViT model.

Additionally, we calculated the forward GFLOPs for models with the same $H / FF$ shown in Table \ref{table:vision_imagenet}, namely ViT, OWF, UiT and RingFormer. For an RGB input image of size 224x224 (applied with 16x16 patch size), ViT, UiT and OWF models exhibit similar computational costs of around 17.636 GFLOPs, while RingFormer requires slightly more computations at 19.03 GFLOPs. This increase is attributed to the additional depth-specific and input-dependent level signals used in RingFormer to improve performance while maintaining a lower parameter count compared to standard Transformers.

\begin{figure*}[h!]
    \centering
    \begin{minipage}{0.25\textwidth}
        \includegraphics[width=\textwidth]{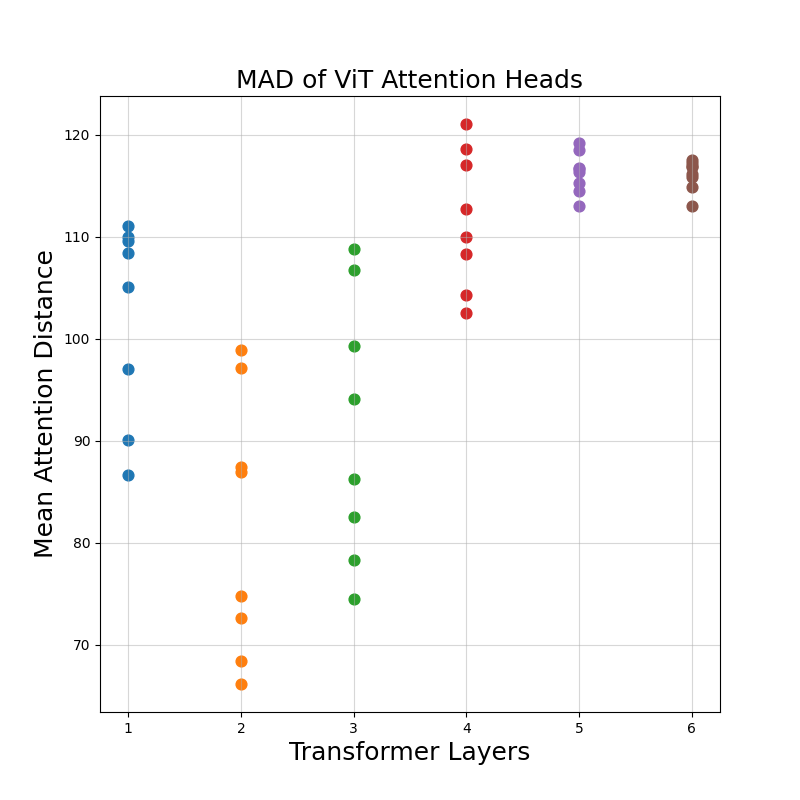}
    \end{minipage}%
    \hfill
    \begin{minipage}{0.25\textwidth}
        \includegraphics[width=\textwidth]{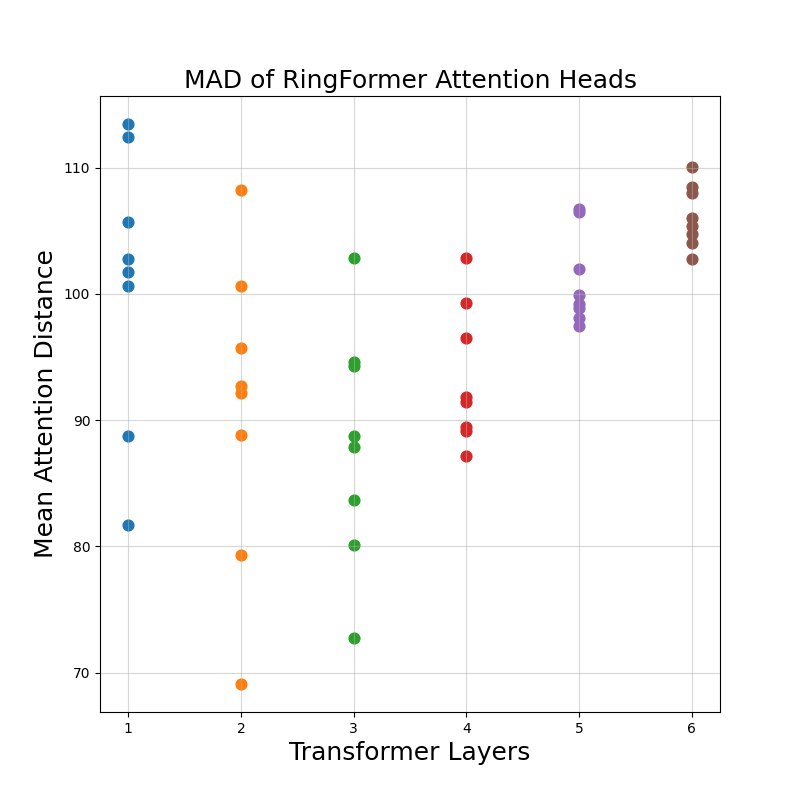}
    \end{minipage}%
    \hfill
    \begin{minipage}{0.25\textwidth}
        \includegraphics[width=\textwidth]{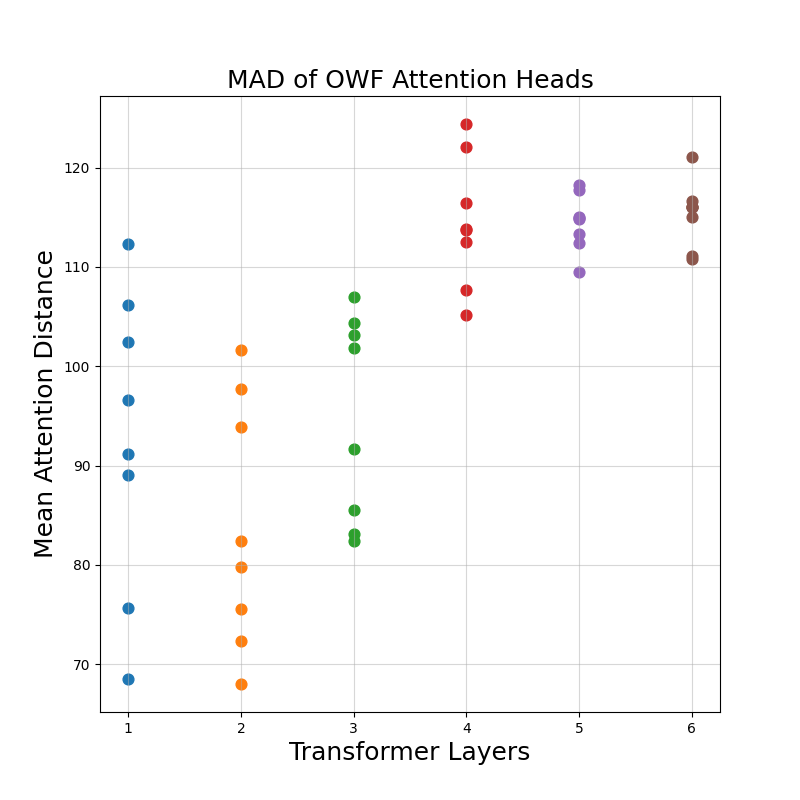}
    \end{minipage}%
    \hfill
    \begin{minipage}{0.25\textwidth}
        \includegraphics[width=\textwidth]{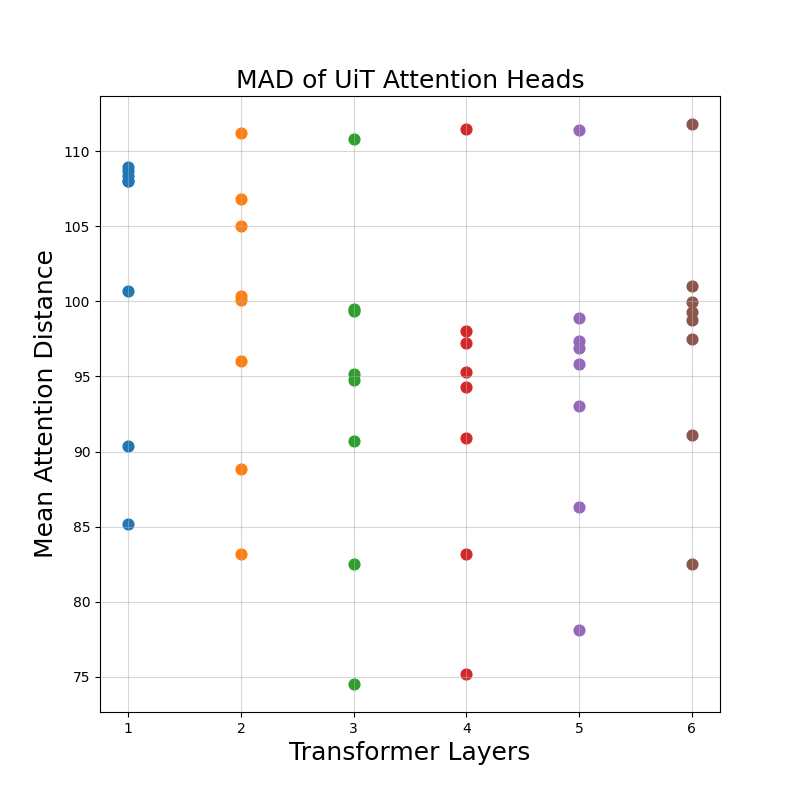}
    \end{minipage}
    \begin{minipage}{0.25\textwidth}
        \includegraphics[width=\textwidth]{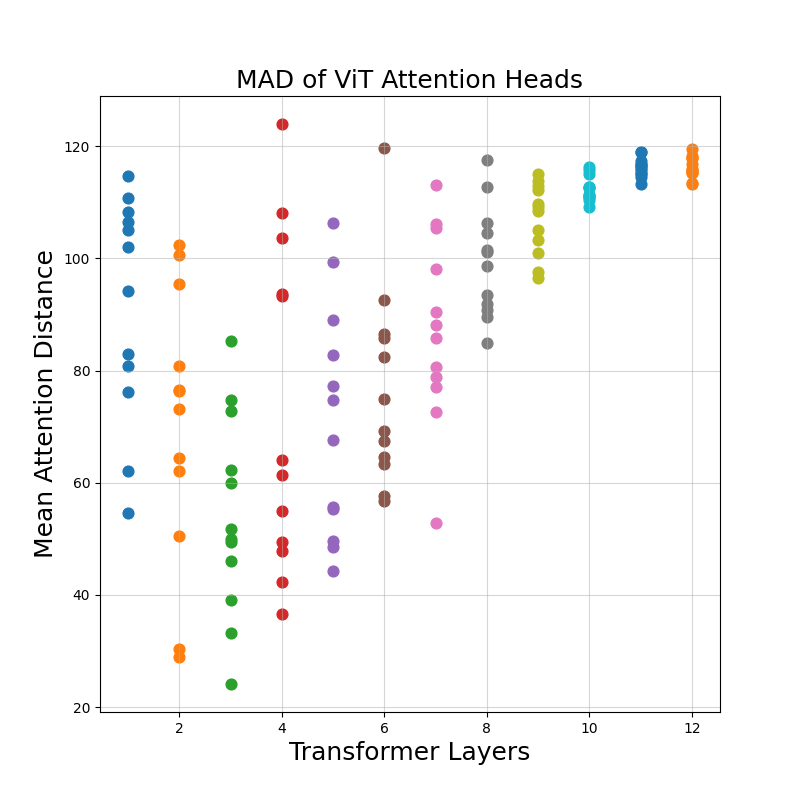}
    \end{minipage}%
    \begin{minipage}{0.25\textwidth}
        \includegraphics[width=\textwidth]{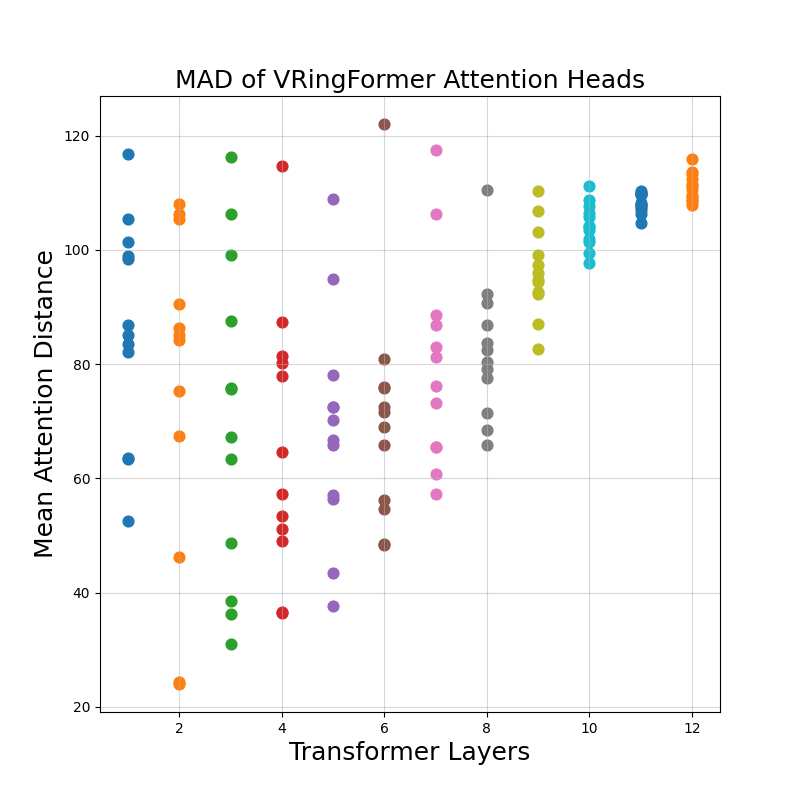}
    \end{minipage}%
    \begin{minipage}{0.25\textwidth}
        \includegraphics[width=\textwidth]{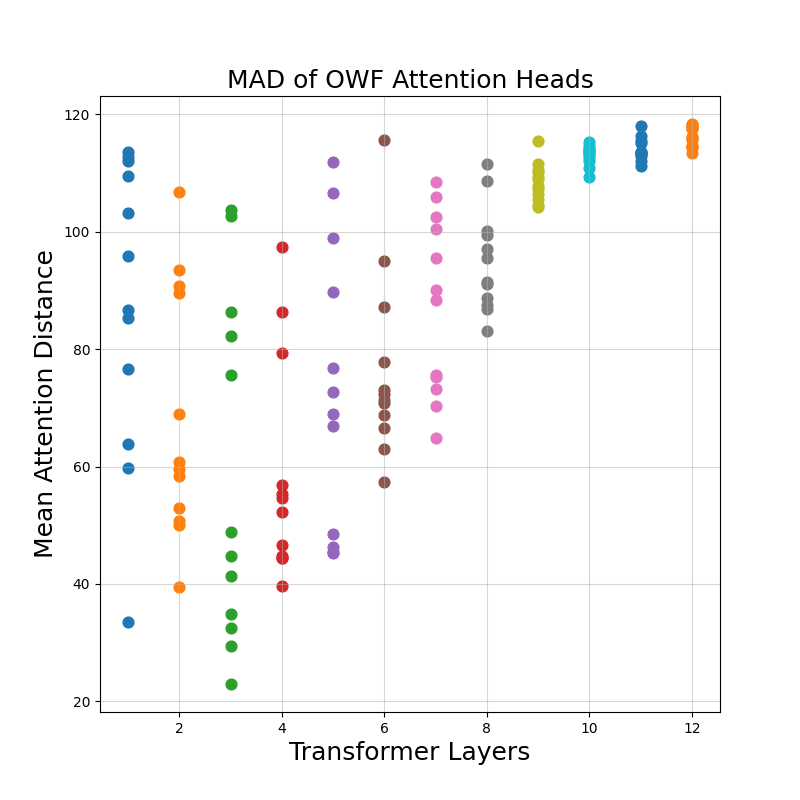}
    \end{minipage}%
    \begin{minipage}{0.25\textwidth}
        \includegraphics[width=\textwidth]{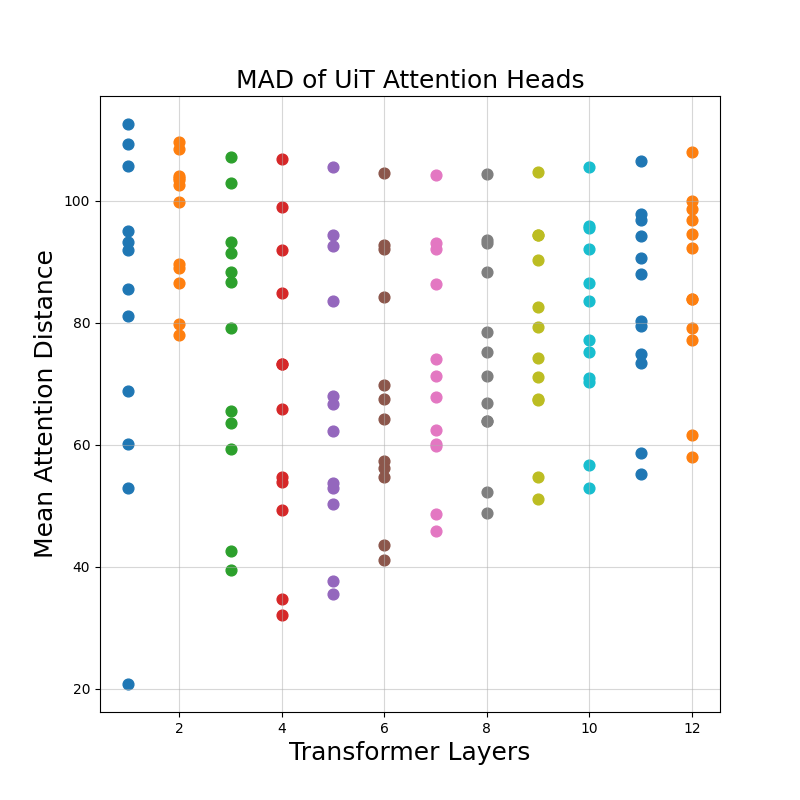}
    \end{minipage}
    \caption{MAD (mean attention distance) analysis for the models trained on image classification task: ViT, RingFormer, OWF - One Wide FFN, UiT - Universal Transformer. The smaller models shown on the upper row have 8 attention heads, and larger models shown on the lower row have 12 attention heads. The points on the plots represent the mean attention distance of an attention head belonging to a particular Transformer layer.}
    \label{fig:grid_mda}
\end{figure*}

\paragraph{Representation Similarity Analysis} To analyze representations across layers / iterations between the original Transformer model and other models, we utilized CKA \cite{kornblith2019similarity} method as shown in Figure \ref{fig:grid_cka_base}. We performed this analysis on base size models, for which we used 3K test source-target pair of sentences from WMT-14 De-En \cite{bojar-etal-2014-findings}. The similarity scores on the diagonal axis in the sub-figures indicate how close the layers (sharing the same index) are between models. We found that RingFormer closely matches the Vanilla Transformer \cite{vaswani2017attention} along with One Wide FFN \cite{pires2023wide}, while Universal Transformer \cite{dehghani2019universal} shows lower similarity. We also report the analysis results for large models in Appendix \ref{appendix:additional_analysis}.

\paragraph{Mean Attention Distance Analysis} To study the qualities of attention heads in the vision models, we perform MAD analysis, which is conducted in the original ViT paper \cite{dosovitskiy2020image}. We first do the analysis on the smaller models trained on \textit{ImageNet-small} (ViT, UiT$^s$, OWF, and RingFormer$^s$ shown in Table \ref{table:vision_mini_imagenet}), and also on the larger size models trained on \textit{ImageNet-1K} (ViT, OWF, UiT$^s$ and RingFormer$^s$ in Table \ref{table:vision_imagenet}). We computed mean attention distances of 500 images randomly taken from the \textit{ImageNet-small} validation set and took their average. The MAD analysis plots for each model above are shown in Figure \ref{fig:grid_mda}.

We observe that, in the ViT model, different attention heads yield different attention distances suggesting they use both local and global information from an image. But as we go deeper in the Transformer blocks, the heads tend to focus more on global aggregate information. The same type of phenomenon occurs in the case of One Wide FFN model, which is expected as its attention layers are not recurrent. In the case of our RingFormer model, the properties of its attention heads are also very similar to those of the ViT model. It can be seen as a validation of our hypothesis that the level signals could successfully steer the behavior of a recurrent Transformer model as it goes through a series of iterations. When it comes to the Universal Transformer model, it is found that the types of signals that exist in that model could not sufficiently help it simulate its attention module behavior as it is considerably different compared to that of the ViT model.

\subsection{Ablation Study}
\label{ablations}
In this section, we conduct an ablation study with various experiments on the translation task to validate the effectiveness of our proposed method. The training details for ablation study are the same as those of main translation models, except smaller batch size (128), hidden input dimension and feedforward block dimension. The model hyperparameters and experimental results for these ablation studies are presented in Table \ref{table:ablation}.

 First, we train a recurrent Transformer using static level signals introduced in Universal Transformers \cite{dehghani2019universal}, which has the lowest performance. When we drop either attention level signals or FFN level signals, ``w.o. attn'' and ``w.o. FF'' in Table \ref{table:ablation}, the performance degradation occurs compared with other variations where those signals are present.
 Also, we do the following two ablations: 1) we add level signals ``before attn'' projection while keeping our original design for FF level signals, 2) we add level signals, ``inter-FF signal'', after intermediate feedforward projection like $FFN(x) = \sigma(x W_{up} + g_{Fi}(x))W_{down}$, while keeping our original design for attention level signals. The performances of those two experiments are almost the same but lower than our design choice where additions occur i) after attention projection and ii) before the up-projection layer of the FF block. In addition, when we use smaller rank, H / 32, compared to our default rank, H / 16, the performance decreases, but when we increase the rank or make the matrix full-rank to generate level signals, as expected, the models show better performance.

\begin{table}[h!]
\footnotesize
\centering
\begin{tabular}{c|c c|c}
\toprule
{$model$} & {$H / F$} & {$P$} & {$BLEU\uparrow$} \\
\midrule
static signal & 128 / 512 & 20.48M & 23.35 \\
%w.o attn \& FF signal & 128 & 512 & 20.48M & 7.62 \\
w.o. attn signal & 128 / 512 & 20.51M & 24.23 \\
w.o. FF signal & 128 / 512 & 20.59M & 24.37 \\
before attn & 128 / 512 & 20.58M & 24.56 \\
inter-FF signal & 128 / 512 & 20.56M & 24.58 \\
\midrule
H / 32 rank signal & 128 / 512 & 20.53M & 24.21 \\
H / 8 rank signal & 128 / 512 & 20.68M & 24.96 \\
full-rank signal & 128 / 512 & 21.27M & 25.37 \\
\midrule
\textbf{Ours} & 128 / 512 & 20.58M & 24.92  \\
\bottomrule
\end{tabular}
\caption{Ablation experiment results of translation task in WMT-14 \cite{bojar-etal-2014-findings} German-English pairs with various model-designs. Each model is evaluated by BLEU \cite{papineni2002bleu} score on the test set. The $H$, $FF$, and $P$ represent the hidden input dimension, feedforward block dimension, and total parameter size, respectively.}
\label{table:ablation}
\end{table}

\section{Conclusion}
\label{Conclusion}
In this paper, we introduce \textit{RingFormer}, a parameter-efficient recurrent Transformer architecture that employs a single Transformer layer recurrently while integrating input-dependent signal vectors created using low-rank matrices for each level. This approach significantly reduces the number of parameters while maintaining high performance in tasks such as machine translation and image classification. We hope that our research on enhancing recurrent Transformer with adaptive level signals can enable smaller organizations and research institutions to train powerful models without the need for extensive computational resources, thus democratizing access to advanced AI capabilities.

\section{Limitations}
Our approach introduces additional computations compared to the original Transformer due to the integration of depth-specific and input-dependent signals. However, this trade-off is necessary to maintain the performance of standard Transformers while significantly reducing parameter count compared to other recurrent Transformer models. Due to computational constraints, we were not able to conduct experiments on large-scale language modeling tasks, which require significantly more data and training resources, and our experiments were limited to relatively smaller scale models. While our design choices suggest that RingFormer should retain its advantages at larger scales, future work can focus on further validating its performance on billion-parameter models and explore its effectiveness in domains such as language modeling.
\label{limitations}

\bibliography{custom}

\appendix
\section{Appendix}
\label{sec:appendix}

\begin{figure*}[h]
    \begin{minipage}{0.3\textwidth}
        \includegraphics[width=\textwidth]{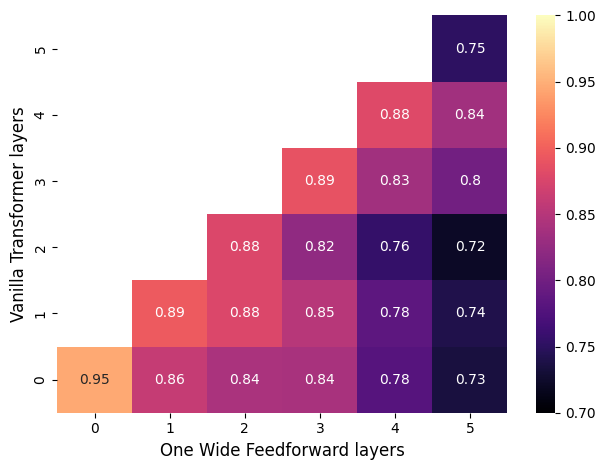}
    \end{minipage}%
    \hfill
    \begin{minipage}{0.3\textwidth}
        \includegraphics[width=\textwidth]{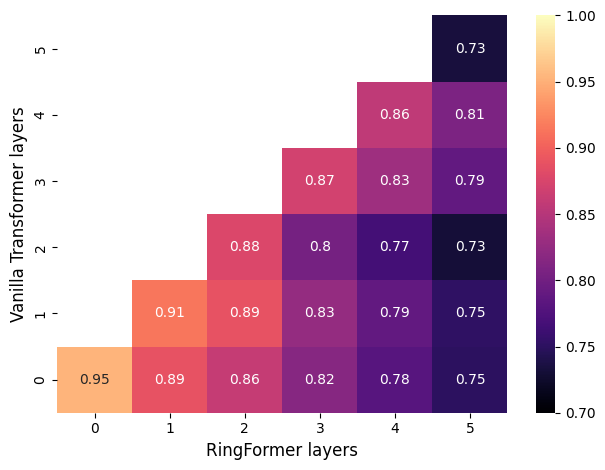}
    \end{minipage}%
    \hfill
    \begin{minipage}{0.3\textwidth}
        \includegraphics[width=\textwidth]{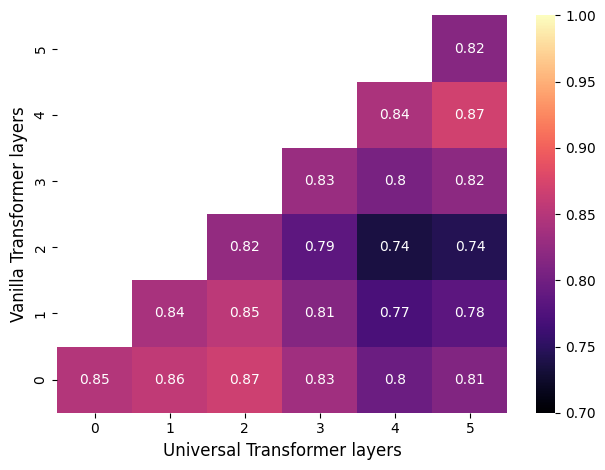}
    \end{minipage}%
    \vfill
    \begin{minipage}{0.3\textwidth}
        \includegraphics[width=\textwidth]{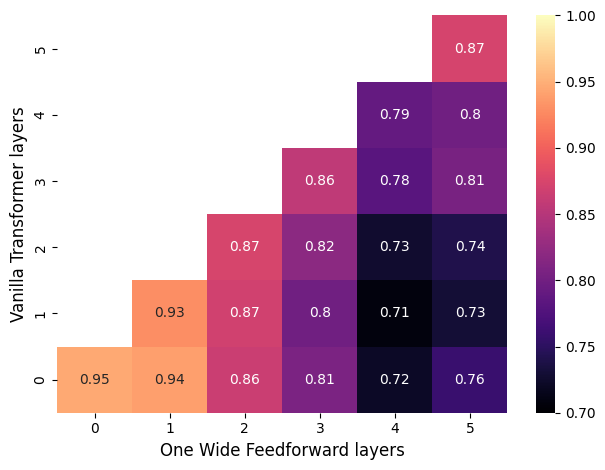}
    \end{minipage}
    \hfill
    \begin{minipage}{0.3\textwidth}
        \includegraphics[width=\textwidth]{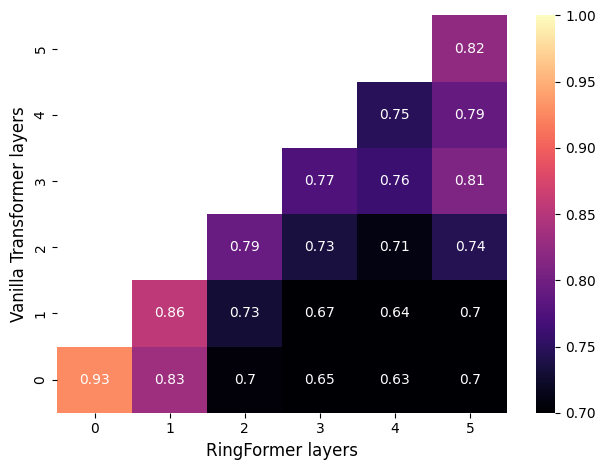}
    \end{minipage}
    \hfill
    \begin{minipage}{0.3\textwidth}
        \includegraphics[width=\textwidth]{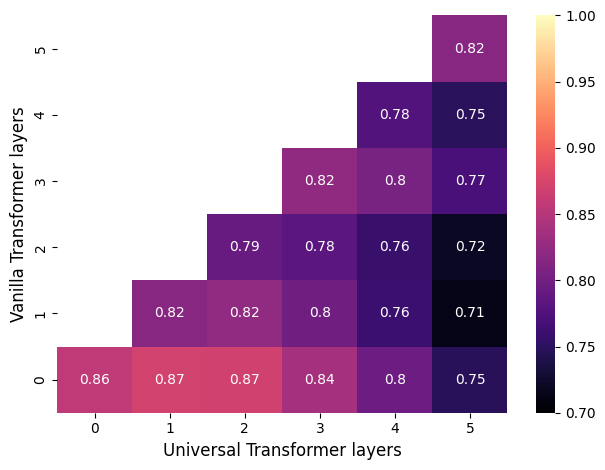}
    \end{minipage}
    \caption{Representation Similarity Analysis using CKA (centered kernel alignment) \cite{kornblith2019similarity} for the large-size models trained on the translation task: Transformer, Ring - RingFormer, OWF - One Wide FFN, Uni - Universal Transformer.  The figures on the upper row are for the encoder part. The figures on the lower row means are for the decoder part. All models have 6 number of layers / iterations. The values on the figures are between 0 and 1, where higher values indicate more similarity of layers between models.}
    \label{fig:grid_cka}
\end{figure*}

\subsection{Implementation Details}
\label{appendix:implementation_details}

\paragraph{Translation}
Models are trained based on the two size variations, base size and large size. The base size models are trained based on the following model configuration settings: 6 Transformer layers, 8 attention heads, 512 hidden dimension size, 2048 feedforward dimension with maximum sequence length 50. For training, their maximum learning rate is 7e-4 with 17K step cosine warm-up scheduler and total 210K training steps on two A100 80GB GPUs. The large size models are trained based on the following model configuration settings: 6 Transformer layers, 16 attention heads, 1024 hidden dimension size, 4096 feedforward dimension with maximum sequence length 50. For training, their maximum learning rate is 2e-4 with 17K step cosine warm-up scheduler and total 210K training steps on two A100 80GB GPUs.

\paragraph{Image Classification} For the models trained on \textit{ImageNet-small} dataset, we used 224x224 image resolution, 16x16 patch size, 6 Transformer layers, 8 attention heads, learning rate of $1e^{-3}$, cosine learning rate scheduler with 2K warm-up steps, batch size of 1024, and training for 9775 steps (100 epochs) with one RTX 3090 GPU. For the models trained on \textit{ImageNet-1K} dataset, we used the same image resolution and patch size as mentioned above, 12 Transformer layers, 12 attention heads, learning rate of $5e^{-4}$, cosine learning rate scheduler with 3128 warm-up steps (5 epochs), batch size of 4096, 16 gradient accumulation steps, and training for around 15650 steps (50 epochs) on two RTX 3090 GPUs.\\

For all models, we used dropout rate of 0.1, gradient clipping of 1.0 during training and GELU \cite{hendrycks2023gaussian} activation function.

\subsection{Additional Analysis}
\label{appendix:additional_analysis}
In Figure \ref{fig:grid_cka}, we share the representation similarity analysis for big size models in the Translation task. This analysis also has been conducted under the same conditions as in the base size case. Similar with the results in Figure \ref{fig:grid_cka_base}, One Wide FFN \cite{pires2023wide} and our RingFormer model have higher layer-wise representations with the Vanilla Transformer \cite{vaswani2017attention} compared to Universal Transformer \cite{dehghani2019universal}.

\subsection{ImageNet-small Dataset}
\label{appendix:imagenet_small_dataset}
We sampled a subset of \textit{ImageNet-1K} \cite{deng2009imagenet} that contains randomly selected 100 classes, with 100,000 images for training and 5000 images for testing, in order to perform experiments on smaller size models. In the project Github repository, we will share the names of all the sampled images for training and testing as a json file. \\

\end{document}